 \documentclass[pmlr,twocolumn,10pt]{jmlr} 

\usepackage{outlines}
\usepackage[T1]{fontenc}
\usepackage{multirow}
\usepackage{graphicx}
\usepackage{booktabs}
\usepackage{outlines}
\usepackage{siunitx}

\usepackage[switch]{lineno}

\newcommand{\equal}[1]{{\hypersetup{linkcolor=black}\thanks{#1}}}
\newcommand{\corresp}[1]{{\hypersetup{linkcolor=black}\thanks{#1}}}

\usepackage{csquotes}
\usepackage{listings}
\usepackage{amsmath}
\usepackage{amssymb,amsmath}
\usepackage{wrapfig}
\usepackage{tcolorbox}
\usepackage{lipsum} 

\definecolor{darkgreen}{rgb}{0.0, 0.5, 0.0}

\definecolor{codegreen}{rgb}{0,0.6,0}
\definecolor{codegray}{rgb}{0.5,0.5,0.5}
\definecolor{codepurple}{rgb}{0.58,0,0.82}
\definecolor{backcolour}{rgb}{0.95,0.95,0.92}

\lstdefinelanguage{promptlanguage}{
    morecomment=[l][\color{codepurple}]{//},
    morecomment=[s][\color{blue}]{\{\{}{\}\}},
}
\lstdefinestyle{promptstyle}{
    backgroundcolor=\color{white},   
    commentstyle=\color{codegreen},
    keywordstyle=\color{magenta},
    numberstyle=\tiny\color{codegray},
    stringstyle=\color{codepurple},
    basicstyle=\ttfamily\small,
    frame = single,
    breakatwhitespace=false,         
    breaklines=true,                 
    captionpos=b,                    
    keepspaces=true,                 
    numbers=left,          
            xleftmargin=0.5cm,
        xrightmargin=0.5cm,
    numbersep=5pt,                  
    showspaces=false,                
    showstringspaces=false,
    showtabs=false,                  
    tabsize=2
}
\makeatletter\lst@AddToHook{PreSet}{\exhyphenpenalty=50 }\makeatother

\newenvironment{chat} {
    
    \newcommand\who[1]{\par\hangindent=2.5em\hangafter=1 ##1:}
    \par\vskip2em
}{%
    \par
    \hangindent=0pt\hangafter=0
    \vskip2em
}

\newcommand{\methodname}{MED-OMIT\xspace}

\theorembodyfont{\upshape}
\theoremheaderfont{\scshape}
\theorempostheader{:}
\theoremsep{\newline}

\renewcommand{\cite}[1]{\citep{#1}}

\jmlrvolume{259}
\jmlryear{2024}
\jmlrsubmitted{LEAVE UNSET}
\jmlrpublished{LEAVE UNSET}
\jmlrworkshop{Machine Learning for Health (ML4H) 2024} 

 \title[MED-OMIT]{\methodname: Extrinsically-Focused Evaluation Metric for Omissions in Medical Summarization}

\author{%
\centering
\Name{Elliot Schumacher}\corresp{\texttt{elliot@curai.com}} ~~
\Name{Daniel Rosenthal} ~~
\Name{Dhruv Naik} ~~
\Name{Varun Nair}\equal{Work done while at Curai Health} \\
\Name{Luladay Price}\footnotemark[2] ~~
\Name{Geoffrey Tso}\footnotemark[2] ~~
\Name{Anitha Kannan}\\
\addr \centerline{Curai Health}
 }

\begin{document}

\maketitle

\begin{abstract}

Large language models (LLMs) have shown promise in safety-critical applications such as healthcare, yet the ability to quantify performance has lagged. An example of this challenge is in evaluating a summary of the patient's medical record. A resulting summary can enable the provider to get a high-level overview of the patient's health status quickly.  Yet, a summary that omits important facts about the patient's record can produce a misleading picture. This can lead to negative consequences on medical decision-making.

We propose \methodname as a metric to explore this challenge.  We focus on using provider-patient history conversations to generate a subjective (a summary of the patient's history) as a case study.  We begin by discretizing facts from the dialogue and identifying which are omitted from the subjective.  To determine which facts are clinically relevant, we measure the importance of each fact to a simulated differential diagnosis. 
We compare \methodname's performance to that of clinical experts and find broad agreement
We use \methodname to evaluate LLM performance on subjective generation and find some LLMs (gpt-4 and llama-3.1-405b) work well with little effort, while others (e.g. Llama 2) perform worse.

\end{abstract}

\paragraph*{Data and Code Availability}
We use the Ambient Clinical Intelligence Benchmark (ACI-BENCH) corpus \cite{aci-bench}, which is openly available.   We will release the code at \url{https://github.com/curai/curai-research/tree/main/MEDOMIT}.

\paragraph*{Institutional Review Board (IRB)}
This research does not require an Institutional Review Board as the datasets used in this work are publically available datasets.

\linespread{1}

\begin{figure*}[tb]
\begin{tcolorbox}[boxsep=1pt,left=2pt,right=2pt,top=0pt,bottom=0pt]
\textbf{\footnotesize Subjective}

\small
Chief Complaint: The patient, Stephanie, a 49-year-old female, has been experiencing increased fatigue and dizziness over the past couple of months. She reports feeling worn out from daily activities that she used to handle without issue.

History of Present Illness: Stephanie's symptoms have been ongoing for a few months. She has not noticed any blood in her stools, nor have they been dark or tarry. She denies heavy menstrual bleeding, weight loss, loss of appetite, or fainting. She has been feeling dizzy but has not passed out. She has had some nasal congestion due to seasonal allergies.

Past Medical History: Stephanie has a significant past medical history of congestive heart failure, kidney stones, and a colonoscopy due to blood in her stools three years ago. The colonoscopy revealed mild diverticulosis, but she has had no issues since then. She has been struggling with her salt intake due to her congestive heart failure, and she admits to not doing well recently due to travel and eating fast food. She has noticed some swelling in her legs but nothing extreme. She has not had any recent flare-ups of her kidney stones, back pain, or blood in her urine.
\end{tcolorbox}

\begin{tcolorbox}[boxsep=1pt,left=2pt,right=2pt,top=0pt,bottom=0pt]
\small
\textbf{\footnotesize Omitted Facts}

Stephanie has slightly reduced heart function

\quad The summary does not mention Stephanie's slightly reduced heart function (Score: 0.5).

Stephanie went to Vermont to explore the mountains

\quad The summary does not mention Stephanie's recent travel to Vermont (Score: 0.1).
  
Stephanie ate two cheeseburgers at McDonald's during her travel
  
\quad The summary does not mention Stephanie's specific food intake during her travel (Score: 0.1).
  
Stephanie has not experienced any shortness of breath or problems lying flat at
    night
    
\quad The summary does not mention Stephanie's lack of shortness of breath or problems lying flat at night (0.1).
    
Stephanie's hemoglobin is low
  
\quad The summary does not mention Stephanie's low hemoglobin levels (Score: 1).

\end{tcolorbox}
\caption{Example GPT-4 generated subjective paired with the list of omitted facts and their weight.  The facts are generated from the original patient-provider dialogue and their importance is scored using the \methodname pipeline. See Appendix Figures \ref{fig:example_short_chat}, \ref{fig:example_pt1} and \ref{fig:example_pt2} for additional context.}
\label{fig:intro_example}
\end{figure*}

\section{Introduction}
Medical providers face perpetual challenges in maintaining patient documentation \cite{payne2015report, arndt2017tethered}. Automating this work has been made increasingly feasible by large language model (LLMs) \cite{openai2023gpt4,chowdhery2022palm,touvron2023llama,jiang2023mistral}, as LLM-powered note generation has shown an increase in performance compared to previous methods \cite{nair2023dera}. Yet automatically generated clinical notes are imperfect \cite{ben-abacha-etal-2023-investigation}, creating negative consequences for healthcare. 

Issues range from omissions, in which important information is incorrectly excluded from the summary, and hallucinations, in which information is fabricated and included.  Hallucinations are objective and can be detected using comparisons against the original document or external sources \cite{min2023factscore,umapathi2023medhalt,vu2023freshllms,ji2023rho,cohen2023lm,peng2023check}. Yet detecting erroneous omissions is comparatively challenging as they are matters of judgment.

We focus on omissions in the \textit{subjective} section of the clinical note within the SOAP framework \cite{podder2022soap}.  A subjective is a summary of everything relevant to the patient's current health issue and informs the provider how to assess the patient's condition and design a treatment plan.  The provider often uses the subjective summary to determine a differential diagnosis (DDx), a list of possible diagnoses.  As a result, the subjective must contain all potentially relevant information.

\begin{figure*}
    \centering
    \includegraphics[width=0.95\textwidth]{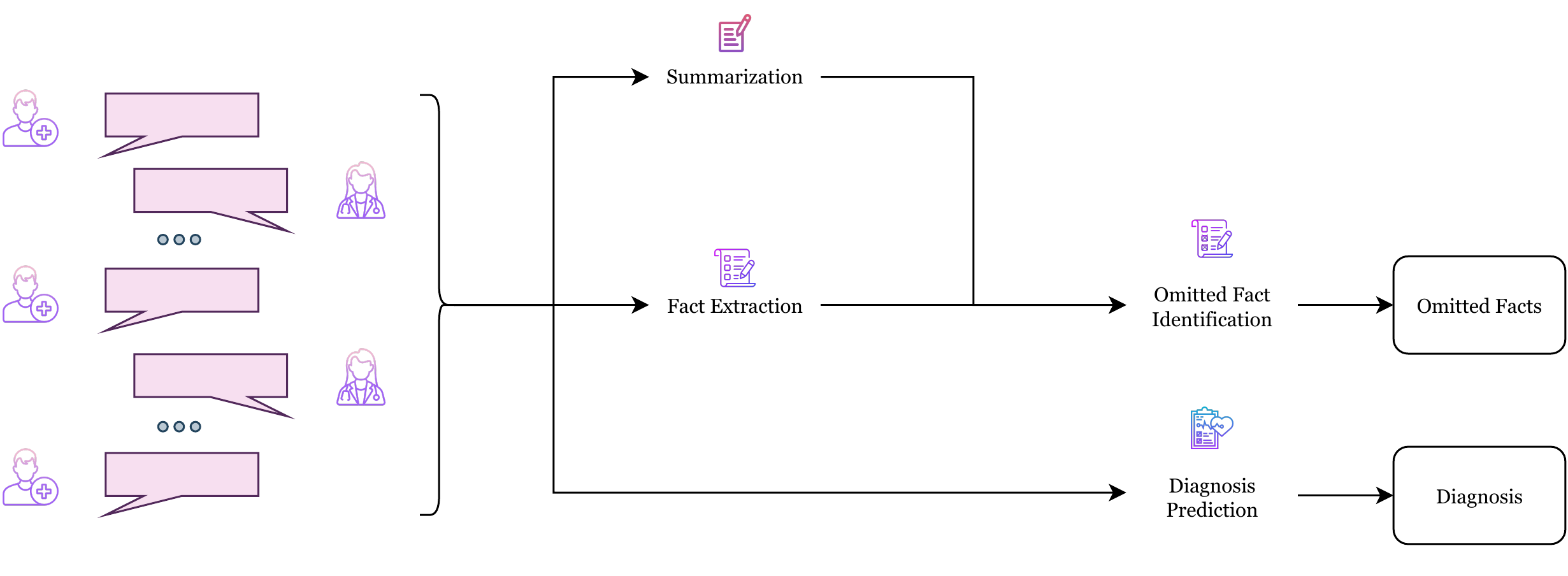}
    \caption{Given a patient-provider dialogue (left), we compute a summary and use a \textit{fact extraction} module to extract facts from the conversation. We use the extracted facts from the conversation to identify if any facts are omitted from the summary. We also compute a differential diagnosis using the conversation data.}
    \label{fig:pipeline_1}
\end{figure*}

\begin{figure*}[tb]
    \centering
    \includegraphics[width=0.9\textwidth]{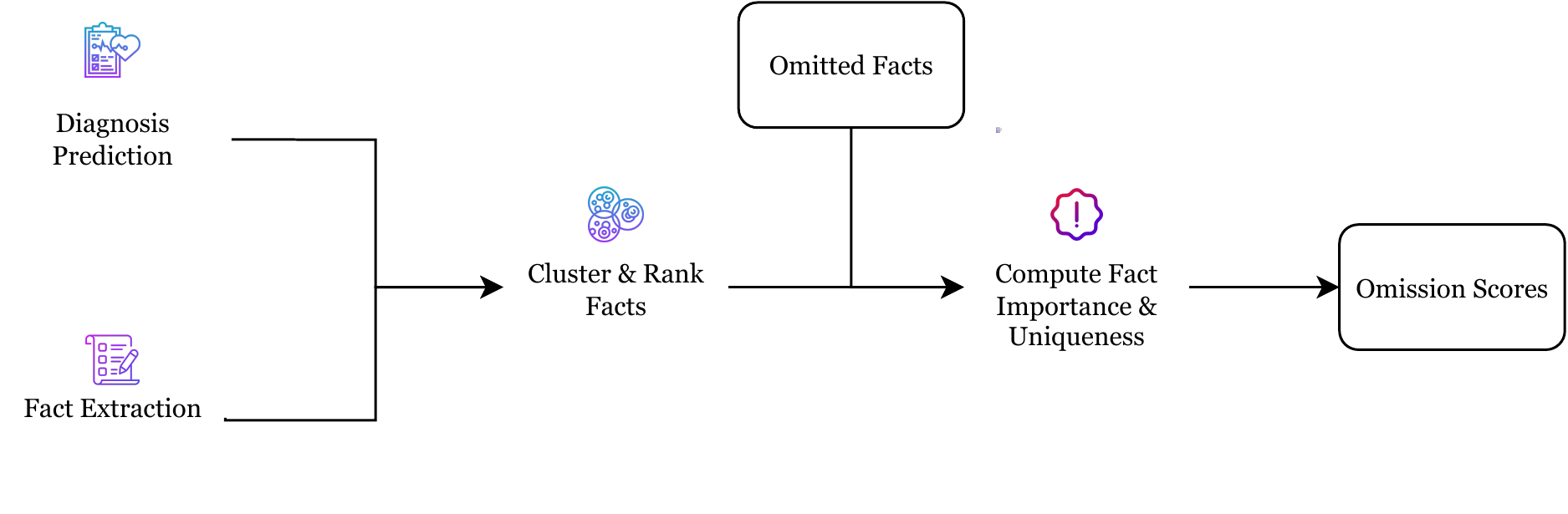}
    \caption{Given the previous outputs of the diagnosis prediction and fact extraction modules, we cluster facts that either support or refute a diagnosis. We also categorize each fact w.r.t. each diagnosis. With the clustered \& categorized facts and the previously computed fact omissions, we assign an importance and uniqueness score to each fact.}
    \label{fig:pipeline_2}
\end{figure*}

Merely detecting which facts are omitted from a subjective insufficiently reflects its quality as irrelevant information \textit{should} be omitted. However,  \textit{important} omissions can mislead a provider.  Therefore, detecting omissions requires identifying omissions \textit{and} quantifying their importance. The importance of an individual fact in a case is multifaceted. Consider the omitted facts in the example in Figure \ref{fig:intro_example}.  A fact such as \textit{Stephanie's hemoglobin is low} is very likely to be relevant to her complaint of fatigue.  Other facts, such as \textit{Stephanie went to Vermont to explore the mountains} are likely less relevant.  Yet, if Lyme disease was potentially suspected, \textit{Stephanie went to Vermont to explore the mountains}, may be critical.  The context is critical in this determination.

We propose \methodname as a multi-step pipeline to produce an omission metric. As shown in Figure \ref{fig:pipeline_1}, We generate a subjective using common LLM-based approaches from the patient-provider chat.  Separately, we generate a list of facts from the conversation, which are atomic pieces of medical information.  Using the list of facts paired with the subjective, we can detect which facts are omitted. 

To identify which facts are important and which are irrelevant, we propose a \textit{fact importance weight} which quantifies the criticality of each omitted fact, illustrated in Figure \ref{fig:pipeline_2}.  We calculate this weight in two ways. First, we do so by categorizing the importance of all facts as a group.  Second, we separately cluster facts that support and refute each diagnosis in an LLM-simulated DDx, and further sub-cluster these by their underlying medical function (or \textit{pathophysiological mechanism}).  This second approach allows us to highlight facts that uniquely point to a diagnosis -- including rare or unlikely ones.  While many facts are highly correlated, this seeks to surface non-correlated facts to the provider even if they are judged unimportant overall.  

Using a simple weight scheme, we generate an importance score for each omitted fact and a cumulative score representing all omitted facts in a subjective.  We compare these metrics against reference-based automated summarization metrics such as BERTScore \cite{zhang2019bertscore} and ROUGE \cite{lin-2004-rouge}. Both BERTScore and ROUGE are designed to be general-purpose metrics and do not target omissions specifically. In an expert annotation analysis, we find that \methodname reflects expert opinion on the presence and importance of each omission.  We find that our reference-free approach reflects the summarization performance of LLMs as they increase in size.  We further find that for larger LLMs, such as gpt-4, there is no correlation between either BertScore or ROUGE and the number of omissions, highlighting the need for a specific-purpose metric.


\section{Background}
Work in large language models, such as gpt-4 \cite{openai2023gpt4}, PaLM \cite{chowdhery2022palm}, Llama \cite{touvron2023llama}, and Mistral \cite{jiang2023mistral}, have enabled advances in text generation performance.  Compared to earlier LLMs such as BERT \cite{devlin-etal-2019-bert}, these model's generations are conditioned on a set of input instructions \cite{reynolds2021prompt, brown2020gpt3}. Summarization tools built on LLMs have shown performance that is equivalent to human-written summaries \cite{zhang2023benchmarking}.  Yet the challenge of quantifying the performance of such approaches has increased as common summarization metrics such as BLEU \cite{papineni-etal-2002-bleu}, ROUGE \cite{zhang2019bertscore}, METEOR \cite{banerjee-lavie-2005-meteor}, and BertScore \cite{zhang2019bertscore} don't align with human judgments \cite{goyal2022news}. Further studies of LLM summarization have also highlighted issues with hallucinations \cite{Ji_2023}.

Therefore, there has been a major focus on developing ways to identify and remediate hallucinations in LLM generations \cite{vu2023freshllms,ji2023rho,cohen2023lm,peng2023check,shuster-etal-2021-retrieval-augmentation,liu-etal-2022-token}.
For example, one work \cite{min2023factscore} proposes to automatically extract atomic facts from the generated text and verify them against an external knowledge source.  In contrast to our work, they weigh each hallucination equally and do not discuss omissions.  In addition, there have been domain-focused hallucination studies in safety-critical domains such as medicine \cite{umapathi2023medhalt}.  Other work has looked at evaluating medical texts using different extrinsic metrics \cite{moramarco-etal-2021-preliminary}.  Relatedly, there is also a line of work that seeks to reduce the risk of harmful LLM output \cite{glaese2022improving, ouyang2022training,scheurer2022training,bai2022constitutional} which is especially important in safety-critical domains such as medicine. To our knowledge, we are unaware of related studies on omission metrics.


\section{Methods}\label{sec:medomit}

What information should and should not be included in a summarization is challenging to determine.
Our metric, \methodname, seeks to quantify this ambiguity through a clinically-motivated approach.  While we believe the insights for this approach can be applied elsewhere, we focus on detecting omissions in subjectives generated from a patient-provider chat.

A subjective note, taken from the SOAP framework \cite{podder2022soap}, consists of the chief complaint (the most pressing medical issue), history of present illness (details about the chief complaint), medical and social history (details about previous medical issues), and current medications and allergies.  To generate a subjective, we adopt the summarization prompt included in \cite{nair-etal-2023-generating}.  The original prompt contains section headers corresponding to the presence, absence, or unknown state of medical findings for the current encounter and medical history.  We altered the section headers to only include information present in the subjective (see Prompt \ref{prompt:summary}).   We focus on using a zero-shot prompt to highlight the model's inherent summarization ability.

Providers often use subjectives to guide the creation of differential diagnoses. 
Mimicking this, we generate a differential diagnosis (DDx) which lists potential medical diagnoses for the patient.  We use the chat as input instead of the summary to provide the most information possible. Separately, we generate a list of facts from the chat, similar to that in \cite{min2023factscore} but medically focused.  This allows us to represent what information is present within the encounter discretely. We can then detect which fact(s) are excluded from the summary.  We define an omission as a fact that is entirely or partially excluded from the resulting summary.  We outline the details of each component in our pipeline.  An example of the output of select pipeline components is included in Appendix Figure \ref{fig:example_pt1} and \ref{fig:example_pt2}.  We also include selected prompts in the Appendix.

\paragraph{DDx}

We prompt the LLM to generate a differential diagnosis given the chat.   This DDx includes at most ten potential medical conditions that might be relevant to the encounter.  Each condition is ranked by order of likelihood, assigned a likelihood category (probable, possible, or unlikely), and given a short explanation.  Note that a patient may have multiple medical issues in a given encounter, so multiple probable conditions may be true.

\paragraph{Fact Identification} We extract a list of facts from the dialogue using a prompt.  This creates a discretized set of facts that is separate from the summary. The prompt is structured to categorize them as medical, related to care access or social determinants of health, or non-medical.  We do not leverage these groups but include them in the prompt to produce high-quality facts. 

\paragraph{Fact Omission Detection} Given the list of facts and the summary, we can then detect which facts are omitted from the summary.  The resulting facts can either be unimportant or very important to clinical decision-making.  However, at this stage, we only make the binary decision of present or omitted.  We adopt a strict definition of a fact being omitted -- if even some portion of the fact (e.g., 'severe' from 'severe pain') is omitted, it is counted as an omission.  We hope future work will explore quantifying the degree of omission.  We create the omission list by using Prompt \ref{prompt:omission_detection}.

\paragraph{Fact Importance Quantification}

At this stage, we have identified a set of facts from the dialogue and which fact(s) are excluded from the summary.  However, the importance of each fact can vary significantly -- a fact such as \textit{The patient has a fever} is likely much more important than \textit{The patient loves iceberg lettuce}. Yet determinations can only be made concerning the specific scenario.  In a different scenario, \textit{The patient loves iceberg lettuce} may be a critical fact if the provider suspects a Listeria infection. Therefore, we employ several approaches to rate the importance of the facts concerning the generated DDx.

First, we assign each fact's importance using three categories, including \textit{critical}, \textit{important}, and \textit{other} (Prompt \ref{prompt:fact_categorization}). We adopt this categorization as a balance between finer-grained methods, such as ranking or scoring each fact individually, and binary categorization.  This determination was made by consulting with a provider and discussing which approach best aligned with their perception of fact importance.

\paragraph{Fact Uniqueness}
Categorizing facts only by their general importance obfuscates other aspects of how a fact might be important. Specifically, facts that uniquely support or refute a specific diagnosis are also critical and may be overlooked with a generic classification approach.   This is especially true for facts that might point to less likely diagnoses, as the previous method is likely to anchor on likely diagnoses. For example, if the only supporting fact for Listeria is \textit{The patient ate iceberg lettuce}, it is important to include it in the subjective even if the DDx determines that Listeria is unlikely.  

Ultimately, the provider should be provided with all evidence for any relevant diagnosis and empowered to make the final determination.  Conversely, in a different scenario, likely multiple correlated facts point to the same underlying symptom (e.g. inflammatory response and fever, headaches, chills).  If one were to be omitted, a clinician could still conclude that the patient had an inflammatory response.

Therefore, we cluster each fact as supporting or refuting evidence concerning each potential diagnosis (e.g., Prompt \ref{prompt:fact_clustering}).  This enables us to create a supporting and refuting evidence list.  For example, \textit{The patient has a fever} would be a supporting fact of a diagnosis of \textit{Influenza}, whereas fever would be inconsistent with \textit{Seasonal Allergies}.

In addition to the first-level clustering approaches, we create sub-clusters for supportive and refuting clusters.  For each group of facts that support a single diagnosis, we prompt the model to cluster facts that suggest the same \textit{pathophysiological mechanism}. This is designed to identify facts that are correlated because they are related to the same underlying issue.  

For example, the facts \textit{Pain at the site of the bursa} and \textit{Swelling at the site of the bursa} both point to potential \textit{Inflammation}. As they are correlated, supporting evidence for inflammation would still be present even if only one fact were included. Yet if a single supporting fact were missing entirely, inflammation would be less likely to be considered.  This intuition leads us to frame the uniqueness as an inverse frequency.  Therefore, a fact's uniqueness would be scored as $\frac{1}{|S|}$, where $S$ is the facts in the subcluster.  See Appendix Figures \ref{fig:example_pt2} and \ref{fig:example_pt3} for examples.

\paragraph{Document-Level Scores}
 
The above section results in a list of omitted facts and their importance.  We further propose a document-level metric for the omitted facts in the summary.  In addition, we explore an alternative metric that seeks to measure the difference in the DDx generated from the chat and the DDx generated from the subjective.

\paragraph{Fact Cumulative Score}
To achieve a document-level score, we individually score each omitted fact by assigning an \textit{importance score} $i$ for each omitted fact.  If the fact omitted was critical, it receives a penalty of 1, a penalty of 0.5 for important, and a penalty of 0.1 for other.  
We separately accumulate a document-level \textit{uniqueness score} $u$.  We assume that facts that uniquely support or contradict a diagnosis are the most important, compared with several facts that point to the same conclusion.  Therefore, we use inverted scoring, where the fact is assigned a score of 
$\frac{1}{|S|}$ for each cluster it is present in.  We take the maximum value of all potential penalties for an overall fact score.  To achieve a fact score for the entire document, we sum all of the individual scores of all omitted facts;
$$
\sum_{f \in \text{omissions}}{max(i_f, u^{0}_f...u^{k}_f))}
$$

This represents a weighted count of the number of omissions in the document.

\begin{figure*}[tb]
    \centering
    \begin{minipage}{0.48\textwidth}
        \centering
        \includegraphics[width=\textwidth]{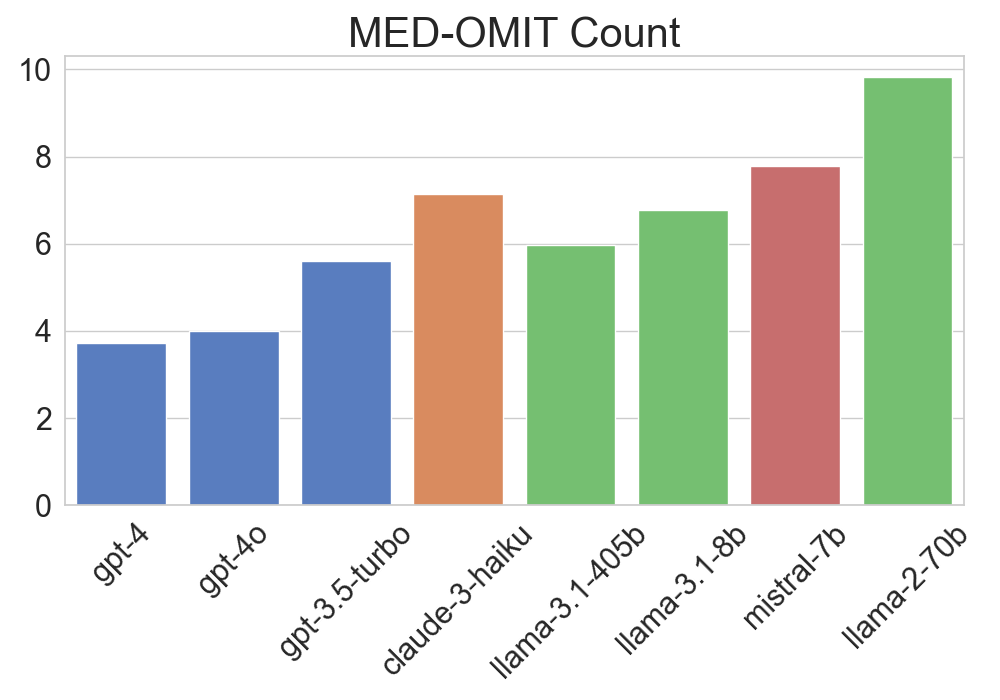}
    \end{minipage}
    \hfill
    \begin{minipage}{0.48\textwidth}
        \centering
        \includegraphics[width=\textwidth]{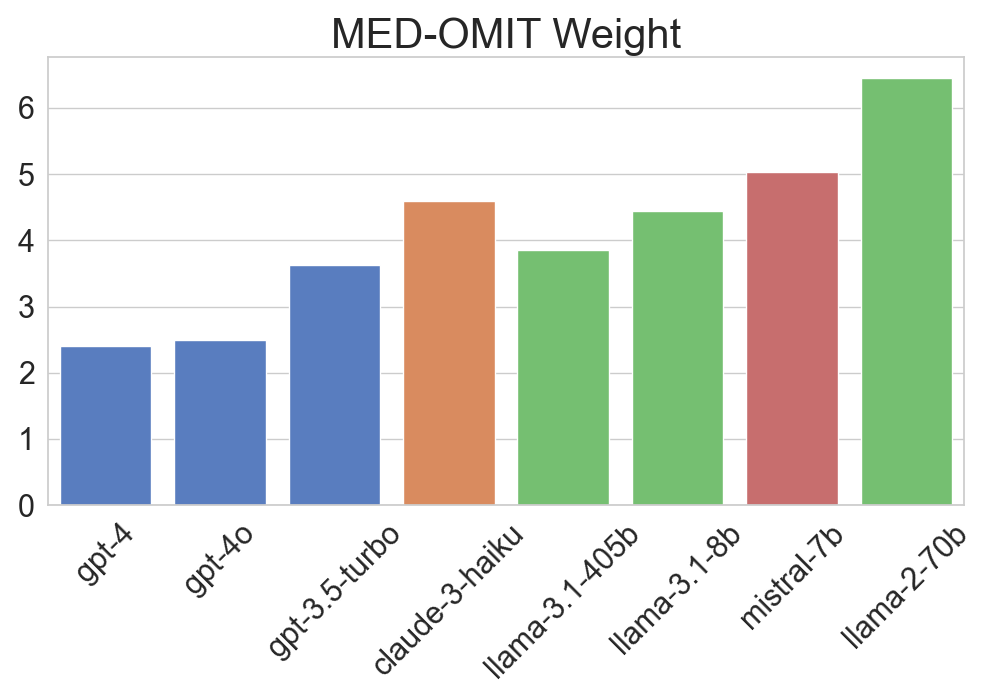}
    \end{minipage}
\caption{For each summary LLM, we calculate the mean of the number of \methodname omissions (left) and the cumulative weight (right), with color indicating model family. A lower score indicates higher performance.  See Appendix Table \ref{tab:weights_and_counts} for full results.}
\label{fig:weights_and_counts}
\end{figure*}


\section{Experimental Setup}

We use the Ambient Clinical Intelligence Benchmark corpus (ACI-BENCH) \cite{aci-bench} to study the efficacy of \methodname. We leverage all three variants of the dataset from this benchmark:  \textit{virtassist} (conversations modeling calls with a virtual assistant), \textit{ virtscribe} (unconstrained directions or discussions with a scribe), and \textit{aci} (natural conversation between a patient and a doctor). We chose to use this dataset for our study as it captures variability in the different forms of conversations that are prevalent today.  Additionally, this allows for replication of our approach which would not be possible with HIPAA-protected medical chats.

We use the training set of 67 chats to calibrate our scoring system and use the three test sets of $118$ chats to evaluate. Two examples from the test set were excluded as their truncated chats were too small to generate a robust subjective.  We truncate the chats using a gpt-4 prompt to exclude non-subjective information (see Appendix \nameref{app:data}). 

\paragraph{Quantitative Experimental Setup}
We separately select which LLM generates a summary and which evaluates the summary.  For the summary prompt (Prompt \ref{prompt:summary}; see the beginning of Section \nameref{sec:medomit}), we select any LLM whose performance we wish to evaluate.  Separately, we can select an LLM for \methodname, which powers the evaluation-focused prompts in Section \nameref{sec:medomit}.  For \textit{summary} models, we evaluate a set of closed-weight models, including gpt-4-0613 (referred to as gpt-4) \cite{openai2023gpt4},  gpt-3.5-turbo, gpt-4o, and claude-3-haiku \cite{claude3}. In addition, we also explore the performance of several open-weight models -- llama-3.1 8 and 405b \cite{dubey2024llama3herdmodels}, llama-2-70B \cite{touvron2023llama}, and mistral-7b \cite{jiang2023mistral}.  For the \methodname model, we use gpt-4-0613 given its higher performance. 
Finally, we also calculate correlation scores with referenced-based metrics BERTScore and ROUGE using the same implementation as used in the dataset paper's code.  As these are referenced-based, we use the ACI gold-standard summaries. Unlike in our generated subjectives, the gold standard notes had access to the entire chat which discussed the final diagnosis.

\paragraph{Medical Expert Evaluation}\label{sec:eval_setup}
In addition, we seek to verify how \methodname's judgments align with those of human clinicians. This is critical not only to judge \methodname's ability to capture LLM performance but also to see if \methodname's incremental judgments, such as cluster creation, are the same as those made by an individual clinician.  Therefore, we ask a group of three medical doctors to validate \methodname.   We focus on our fact omission detection and fact importance approaches for 20 conversations each (60 total).   We randomly selected facts to annotate in each encounter which resulted in 330 fact annotations.  Given the output of  \methodname (using gpt-4 for all prompts), we ask them to answer the following questions. 
\begin{outline}
\1 Was this fact included in the summary? \textit{("Yes", "Partially", "No")}. 
\2  We included the “Partially” option to see how often only a portion of a fact is omitted from the summary. Although we prompt the LLM to make a binary judgment on fact inclusion, there is a continuum between the summary capturing every aspect of the fact and no aspects. 
\1 How many diagnoses are supported by this fact? 
\1 How many diagnoses are refuted by this fact?  
\2 This question and the prior question are simplified forms of the \methodname approach, as we only ask for a count and not the full list of diagnoses.
\1 Finally, if this fact were omitted, how much of an effect would it have on the differential diagnosis? \textit{("Critical", "Important", or "Other")}. 
\end{outline}


\begin{table*}
\centering
\begin{tabular}{c|c|cc|cc}
Fact Missing & Fact Importance & \multicolumn{2}{c}{\# Diagnoses Supp.} & \multicolumn{2}{|c}{\# Diagnoses Ref.} \\
Agreement & Agreement &  Mean Abs. Diff. &  $\sigma$ &  Mean Abs. Diff. & $\sigma$\\ \midrule
80.0\% & 89.3\% & 0.439 & 1.184 & 0.447 & 1.224
\end{tabular}
\caption{Agreement statistics for comparing \methodname using gpt-4 for all prompts with expert annotator decisions on four questions.  For confusion matrices and distribution plots, see Appendix Figures \ref{fig:conf_missing},  \ref{fig:conf_important},  and \ref{fig:ann_diff}. For inner-annotator agreement, see Appendix Table \ref{tab:agreement}.}
\label{tab:annotator}
\end{table*}

\section{Results}

We report \methodname metrics on several Summary - Metric LLM configurations in Figure \ref{fig:weights_and_counts}. We separately report the number of omissions (\methodname Count), and the summation of the omission weights (\methodname Weight). 
For each, we report the mean over the test set  (see Appendix Table \ref{tab:weights_and_counts} for the tabular form which includes standard deviations).  In all metrics, we find that gpt-4 performs best, closely followed by the newer gpt-4o.  The difference between the two gpt-4 versions is likely insignificant and may be due to the model judge being gpt-4.

However, the performance margin between gpt-4 and gpt-3.5-turbo isn't substantial.  It is further remarkable that the \methodname count margin of gpt-3.5-turbo over gpt-4 is larger than that for the \methodname weight, suggesting gpt-3.5-turbo isn't omitting information that is more critical than gpt-4 summaries. The other closed-weight model we evaluate, claude-3-haiku, performs worse than the OpenAI models.

While we find that the open-weight models trail OpenAI models in performance, the gap is narrowing. The gap between the performance of older models (mistral-7b and llama-2-70b) and closed-weight models is quite large. The results of the llama-3.1 models show this gap is narrowing significantly.  Llama-3-405b is competitive with gpt-3.5-turbo, showing major improvements over llama-2. This finding suggests that open-weight models are increasingly viable options for medical tasks.

\subsection{Expert evaluation of \methodname}

As shown in Table \ref{tab:annotator}, we see broad agreement between our medical annotators and \methodname.  First, we find that annotators agree 80\% of the time with MED-OMIT's determination of whether a fact is omitted or not. Second, we find that the agreement on the fact importance question was even higher at 89.3\%.  The confusion matrices in Figures  \ref{fig:conf_missing}  (for fact omission) and Figure \ref{fig:conf_important} (for fact importance) illustrate the results in finer detail and underline the high level of agreement between GPT-4 and our medical annotators.
Additionally, we asked annotators to count the number of diagnoses each fact both supports and refutes.  The absolute difference between the annotator's count and GPT-4's count was less than 0.5 in both cases.  Histograms of the full distributions are available in Appendix Figure \ref{fig:ann_diff}, and illustrate the small amount of disagreement between the annotators and gpt-4 is minor.

Finally, to ensure that the medical experts agreed with each other, we asked each expert to annotate a set of 51 facts distinct from the previous set. As shown in Appendix Table \ref{tab:annotator}, there was broad inter-annotator agreement. In addition to the high exact match rate for the omission and importance questions, we found that Cohen's kappa for each annotator pair showed high to moderate agreement.  There was more disagreement for the supporting and refuting diagnosis counts, but the average maximum difference in results was less than 1 for supportive and less than 2 for refuting counts, which is still reasonable. In summation, these results show that \methodname accurately captures the identifying and quantifying the importance of omissions.

\begin{table*}
\centering
\begin{tabular}{@{}l|ll|cc|cc@{}}
              &                      \multicolumn{2}{c|}{Comparison Metric}      & \multicolumn{2}{c|}{\methodname weight} & \multicolumn{2}{c}{\methodname count} \\
Summary LLM & Name &  Value & spear.         & pear.          & spear.         & pear.         \\ \midrule
gpt-4          &  Rouge LSum    & 0.363       & 0.003            & -0.041           & -0.046           & -0.057          \\
gpt-4          &  BertScore F1 & 0.651       & -0.044           & -0.130           & -0.130           & -0.043          \\
gpt-3.5-turbo &  Rouge LSum    & 0.333       & \textbf{-0.242}  & \textbf{-0.220}  & \textbf{-0.244}  & \textbf{-0.200} \\
gpt-3.5-turbo &  BertScore F1 & 0.627       & \textbf{-0.338}  & \textbf{-0.299}  & \textbf{-0.338}  & \textbf{-0.281}
\end{tabular}
\caption{For the two best models, we compare \methodname mean count and weight to reference-based metrics BERTScore and Rouge. We report the Spearman and Pearson correlation between each reference-based and \methodname metric.  Bolded values are significant with a two-sided test $p<0.05$. For additional metrics, see Appendix Table \ref{tab:app_corr}.}
\label{tab:omm_corr}

\end{table*}


\subsection{Comparison to Traditional Evaluation}

In Table \ref{tab:omm_corr}, we report the Spearman and Pearson correlations between commonly reported summarization metrics (ROUGE and BERTScore) and \methodname (Omission Weight and Counts). Additional metrics are included in Appendix Table \ref{tab:app_corr}. We do not find any significant correlation between the LLM Completion metric and ROUGE or BertScore for larger LLMs such as gpt-4. We find that for the less powerful LLM, traditional summarization metrics correlate slightly with our omission metrics.  Unsurprisingly, higher omission weight and count scores inversely correlate with higher BertScore and ROUGE metrics.  However, there is no statistically significant correlation for summaries generated by more powerful LLMs.

\subsection{Error Analysis}

We performed a qualitative analysis by randomly sampling ten training examples.  While we found \methodname was broadly accurate, there are areas for future improvement.  First, we found that while \methodname was able to consistently detect which facts were omitted from the summary, it did so in a strict manner.  Consider the example in Figure \ref{fig:example_pt1} and \ref{fig:example_pt2} -- a fact (F8) was correctly identified as excluded. However, the summary only omitted the specific foods the patient was excluding from their diet but did include the overall point that she was trying to apply a low-sodium diet.  Capturing the degree of a fact that was excluded remains an open question.

Perhaps the most challenging task is generating the clusters and sub-clusters of supporting and refuting evidence.  Specifically within the framework of the sub-clustering, accurately clustering the facts around symptoms, tests, treatments, and social determinants of health was a challenging prompt to engineer.  While we find that it does well at selecting the correct category and the correct pathophysiological mechanism for the common categories, it can make mistakes.  For example, in Figure \ref{fig:example_pt2}, there is a "NONE" category for symptoms within \textit{Well-managed Congestive Heart Failure}, which is not an actual pathophysiological mechanism.

Additionally, the refuting sub-clustering step occasionally makes broad inferences given the full set of facts.  For example, one refuting sub-cluster noted that \textit{[NAME] has chronic back pain that bothers her when she sits for long periods of time at her desk at work} is a refuting fact for Fibromyalgia because \textit{Fibromyalgia typically presents with widespread pain} even though this is not explicitly stated.  Both LLMs and medical providers make inferences based on what is absent from a medical case, but the amount of alignment is unclear.

Finally, we find that the weighting system does sort summaries pairwise in a sensible manner.  Consider the example in Figure \ref{fig:example_pt2} and another case with only a single omission.   In the single omission case, the fact \textit{Edward experiences swelling in his ankles, mainly near the end of the day} was omitted from a subjective.  This was categorized as critical as it speaks to potential fluid retention which potentially supports several diagnoses.  By contrast, the example in Figure \ref{fig:example_pt2} has five omissions. Yet they are all judged to be less important, and none receive a max score.  This illustrates the importance of going beyond binary judgments on omitted facts.


\section{Conclusion}
We find that \methodname identifies omitted facts and quantifies their importance in line with medical experts. This provides the research community with an important tool in evaluating the capabilities of emerging large language models, and an alternative to small and expensive human evaluations or non-clinically based automated metrics.  The interpretable nature of \methodname can also be used to pinpoint specific problems of omissions subjective generation, guiding where further work is required.

We believe several insights within \methodname generalize to metrics in other medical tasks.  First, discretizing the information present allows for interpretable and meaningful blocks of information.  Identifying whether a fact is or is not included in the summary is much more informative than a similar approach using words alone.  Second, weighing the importance of each fact must be done in line with how a practitioner would do so.  Often, summarization metrics overlook this to create a generalized metric, but in turn, are not useful indicators of performance.


\bibliography{references}
\newpage

\appendix

\newpage
\begin{table*}
\centering

\begin{tabular}{ll|cc|cc}
 &  & \multicolumn{2}{c|}{\methodname Count} & \multicolumn{2}{c}{\methodname Weight} \\
 & Summary LLM & mean & $\sigma$ & mean & $\sigma$ \\ \hline
\multirow{4}{*}{closed} & gpt-4 & 3.72 & 3.07 & 2.41 & 2.31 \\
 & gpt-4o & 4.00 & 3.58 & 2.50 & 2.70 \\
 & gpt-3.5-turbo & 5.61 & 3.56 & 3.63 & 2.57 \\
 & claude-3-haiku & 7.15 & 4.25 & 4.60 & 3.14 \\ \hline 
\multirow{4}{*}{open} & llama-3.1-405b & 5.97 & 4.18 & 3.85 & 3.19 \\
 & llama-3.1-8b & 6.77 & 3.90 & 4.44 & 2.86 \\
 & mistral-7b & 7.79 & 4.21 & 5.03 & 3.01 \\
 & llama-2-70b & 9.82 & 4.33 & 6.45 & 3.19
\end{tabular}
\caption{For each summary  LLM, we calculate the mean and standard deviation of both the number of \methodname omissions and the cumulative weight.}
\label{tab:weights_and_counts}

\end{table*}

\section{Dataset details}\label{app:data}

Our approach is targeted to subjective note, which encapsulates the early part of the encounter where the diagnosis is not necessarily known.
However, the ACI chats discuss the full patient encounter, and include physician-determined diagnoses, outcomes of physical examinations, and test results. Therefore, we truncate the chats to exclude any information that would point to a diagnosis to better simulate when a subjective would be generated.  We find the last relevant line in the chat that discusses any subjective-related information and truncate the chat to the next line using Prompt \ref{prompt:dialogue_truncation}. We will release the truncation indices with our codebase.

\section{Annotation Details}\label{app:ann}

The selected facts consist of all omitted facts in the summary, plus a randomly selected set of facts that were not omitted.  We select all omitted facts and add $n$ more non-omitted facts to annotate at most 5 per encounter.  All values except for the first question were precomputed and presented to the annotator for validation.  The annotators were instructed to change any precomputed value if they believed it appropriate.

The instructions given to the annotators were as follow;
The following sheets contain encounter information from an external dataset.
Each encounter consists of
\begin{outline}
\1 A generated subjective.
\1 A generated differential diagnosis 
\1 A list of all facts extracted from the encounter

\end{outline}

Before answering any questions, please read the above information.

A specific fact from the list is included for consideration. 
With respect to this fact, we'd like you to validate the following questions.  
The values in the first three are pre-computed.  However, you are free to change them if you think appropriate.

\begin{outline}
\1 Is this fact included in the summary?  Rate as No (it is completely excluded), Partially (some element, even a non-medically important one, is excluded), or Yes (it is included)
\1 If this fact is a positive finding, how many diagnosis does it support?  This should be a value between 0 and the total number of diagnoses.
\1 If this fact is a negative finding, how many diagnoses does it refute?  This should be a valute between 0 and the total number of diagnoses.
\1 If this fact were ommitted from the list of facts, what would the impact be on the differential diagnosis?  Please rate as Critical (highest), Important (moderate), Other (lowest).
\2 The impact" of the diagnosis can include a variety of factors.  These include but are not limited to adding a new diagnosis to the list or removing an diagnosis currently present in the list.  Alternatively, would a diagnosis be more or less likely?
\end{outline}

\subsection{Differences between annotators and gpt-4}\label{app:ann_diff}
While there is generally agreement between gpt-4 and annotators, there are several instances where they disagree.  The following are several examples taken from the development data.  We report the fact, the relevant sentence(s) from the summary, and the judgements.

\noindent\rule[0.5ex]{\linewidth}{1pt}

\textbf{Fact:} Vincent experienced dizziness and lightheadedness.

\textbf{Relevant Summary:} He reported experiencing  \\ lightheadedness but denied any noticeable bleeding. 

\textbf{Is Included?:} No (gpt-4, 2 annotators); Partially (1 annotator)

The above example shows the challenge of detecting whether a fact is omitted from the summary.  The summary includes most of the important text, but does exclude \textit{dizziness}.  While related to \textit{lightheadedness}, it is not the same thing.  Since gpt-4 is only allowed to make binary judgements, it says its not included.  Our annotators have the option to select 'Partially'; one decides to do so while the others agree fully with gpt-4.

\noindent\rule[0.5ex]{\linewidth}{1pt}

\textbf{Fact:} Rachel's depression has moments of highs and lows

\textbf{Relevant Summary:} Her depression is managed with Effexor, but she still experiences periods of low mood.

\textbf{Is Included?:} Yes (1 annotator); Partially (2 annotators); No (gpt-4)

This example further illustrates the challenge in determining whether a fact was included.  The majority of the fact is included in the summary.  However, the "highs" work is excluded, which may be informative for the patient's condition.  Since gpt-4 only has a binary choice, it selects No, while the annotators alternatively select Yes or Partially.

\noindent\rule[0.5ex]{\linewidth}{1pt}

\textbf{Relevant Fact:} Mrs. Peterson would avoid going upstairs or downstairs.

\textbf{All facts:} \\
F0: Mrs. Peterson is a 43-year-old patient. \\
F1: Mrs. Peterson is experiencing right leg pain. \\
F2: Mrs. Peterson injured her right leg while bowling.\\
F3: Mrs. Peterson's bowling ball hit her right leg.\\
F4: Mrs. Peterson's right leg has a little bit of bruising on the back end.\\
F5: Mrs. Peterson is able to walk on her right leg, but very carefully.\\
F6: Walking on her right leg is very sore for Mrs. Peterson.\\
F7: Mrs. Peterson would avoid going upstairs or downstairs.\\
F8: Mrs. Peterson has a history of atopic eczema.\\
F9: Mrs. Peterson uses fluocinonide for her eczema when it gets really itchy.\\
F10: Mrs. Peterson has a previous surgical history of a colectomy.\\
F11: Mrs. Peterson had diverticulosis which turned into diverticulitis, leading to the removal of a part of her colon.\\
F12: Mrs. Peterson was bowling when she injured her leg.

\textbf{DDx:}
Contusion (Bruise) :  Probable \\
Muscle Strain :  Probable  \\
Fracture :  Possible \\
Soft Tissue Injury :  Possible \\
Hematoma :  Possible \\
Bursitis :  Unlikely \\
Tendon Rupture :  Unlikely \\
Nerve Damage :  Unlikely \\
Deep Vein Thrombosis (DVT) :  Unlikely \\
Compartment Syndrome :  Unlikely \\
\textbf{Is Important?:} Critical (2 annotator and gpt-4); Important (1 annotator)

While there is less disagreement for fact importance, there are still some tricky cases.  Consider the above case; the fact that the patient is unable to walk up and down stairs should be of obvious concerns to the provider given the hindrance to mobility.  While 2 annotators and gpt-4 decide that it's a critical fact, one annotates it as important.  This is potentially because there are other facts that encapsulate that the patient has trouble walking, and it isn't of strict criticality that she has trouble walking on the stairs.

\begin{figure*}
    \centering
    \includegraphics[width=0.8\textwidth]{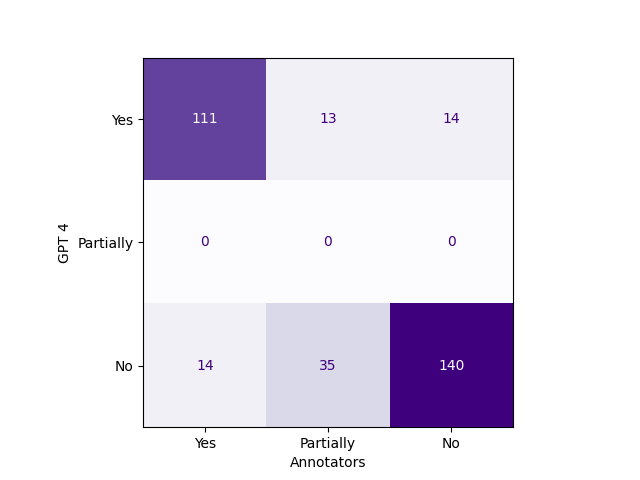}
    \caption{A Confusion Matrix for annotator agreement with GPT-4 for the Fact Omission task. The counts of agreement groups are shown in each cell -- e.g. the number of examples where gpt-4 selected \textit{No}, and annotators selected \textit{Partially} is 35.  The overall agreement was 80\%. Note that while we give annotators three labels to choose from, \methodname only uses a binary judgment (and excludes the "Partially" option). Therefore, we count annotators selecting "Partially" as correct if \methodname selects "Yes").  We believe work capturing the degree of omission would provide further insight.}
    \label{fig:conf_missing}
\end{figure*}
\begin{figure*}
    \centering
    \includegraphics[width=0.8\textwidth]{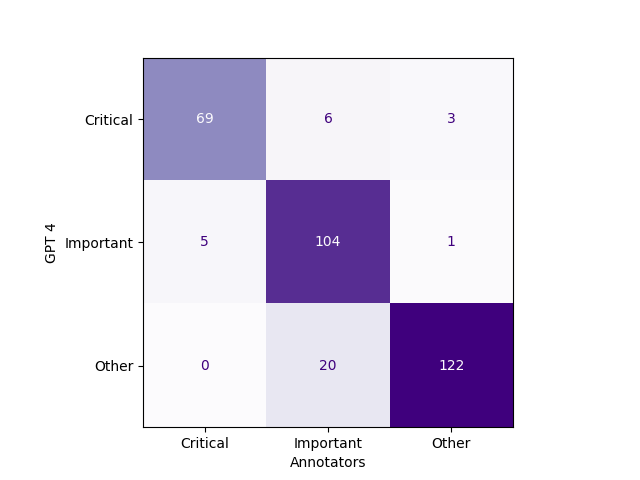}
    \caption{Confusion Matrix for annotator agreement with GPT-4 for the Fact Importance categorization task.  The strict agreement is 89.4\%.}
    \label{fig:conf_important}
\end{figure*}

\begin{figure*}
  \includegraphics[width=0.5\linewidth]{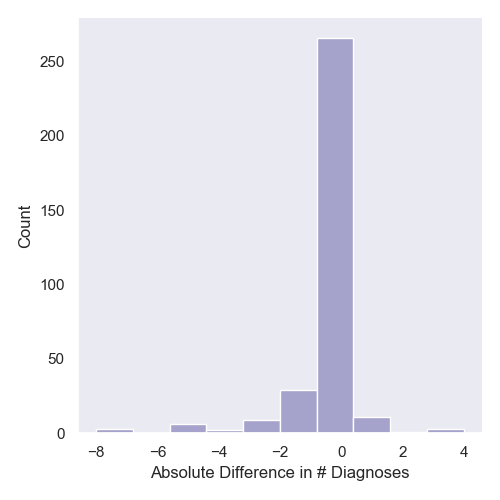}
      \centering

  \caption{Distribution of absolute differences between number of diagnoses supported by each fact as determined by \methodname and expert annotators.}
      \label{fig:ann_diff}

  \end{figure*}

\begin{figure*}
  \includegraphics[width=0.5\linewidth]{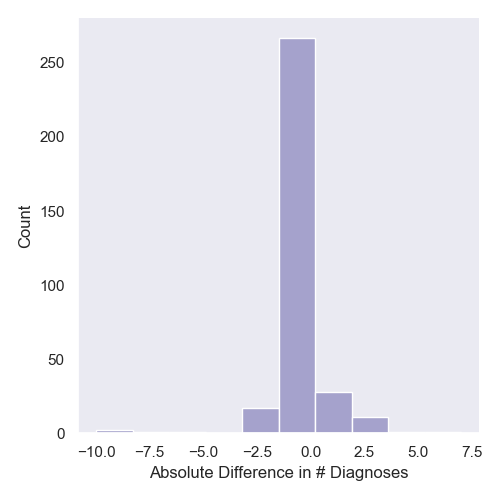}
      \centering

  \caption{Refuting}
    \caption{Distribution of absolute differences between number of diagnoses refuted by each fact as determined by \methodname and expert annotators.}
\end{figure*}

\begin{table*}
\centering
\small
\begin{tabular}{l|lll}
Question & Exact Match & Cohen's Kappa & Cohen's Kappa (Linear) \\ \midrule
Is Included? & 82.35\% & 0.70,0.70,0.74 & 0.76,0.80,0.80 \\
Is important? & 78.43\% & 0.68,0.65,0.62 & 0.74,0.71,0.71 \\ \bottomrule \bottomrule
Question & Exact Match & Max Diff Mean & Max Diff $\sigma$ \\ \midrule
Supports how many diagnosis? & 64.71\% & 0.78 & 1.29 \\
Refutes how many diagnoses? & 58.82\%  &1.84  & 2.80 \\
\end{tabular}
\caption{Inner-annotator agreement statistics for a separate dataset of 51 facts that were annotated by all three annotators.  }
\label{tab:agreement}
\end{table*}


\begin{figure*}[p]
\begin{tcolorbox}[boxsep=1pt,left=2pt,right=2pt,top=0pt,bottom=0pt]
\textbf{\footnotesize Provider-Patient Chat}
\small
\vspace{-2em}
\begin{chat}
\ldots
\who{Provider} hi , stephanie . how are you ?
\who{Patient}  i'm doing okay . how are you ?
\who{Provider}  i'm doing okay . um , so i know the nurse talked to you about dax . i'd like to tell dax a little bit about you , okay ?
\who{Patient}  okay .
\who{Provider}  so , stephanie is a 49-year-old female with a past medical history significant for congestive heart failure , kidney stones and prior colonoscopy who presents today for an abnormal lab finding . so , stephanie , i called you in today because your hemoglobin is low . um , how have you been feeling ?
\who{Patient}  over the past couple of months , i've been really tired and dizzy . lately , i've been really just worn out , even just , you know , walking a mile or going to work , doing things that i've done in the past every day that have been relatively okay , and i have n't gotten tired . and now , i've been getting tired .
\who{Provider}  okay , yeah . i , you know , the nurse told me that you had called with these complaints . and i know that we have ordered some labs on you before the visit . and it did , it c- you know , your , your , your hemoglobin is your red blood cell count . and now , and that came back as a little low on the results , okay ? so , have you noticed any blood in your stools ?
\who{Patient}  uh , no , i have n't . i did about three years ago , um , and i did a colonoscopy for that , but nothing since then .
\who{Provider}  okay , yeah . i remember that , okay . and how about , you know , do your stools look dark or tarry or black or anything like that ?
\who{Patient}  no , nothing like that .
\who{Provider}  okay . and have you been , um , having any heavy menstrual bleeding or anything like that ?
\who{Patient}  no , not that i've noticed .
\who{Provider}  okay , all right . and any , have you passed out at all , or anything like that ? any weight loss ?
\who{Patient}  no , no weight loss or passing out . i have felt a bit dizzy , but it has n't l- led to me passing out at all .
\who{Provider}  okay . so , you endorse some dizziness . you endorse some fatigue . have you , but you have n't had any weight loss , loss of appetite , anything like that ?
\who{Patient}  no , nothing like that .
\who{Provider}  okay , all right . so , you know , let's talk a little bit about that colonoscopy . i know you had a colonoscopy about three years ago and that showed that you had some mild diverticuli- diverticulosis . um , no issues since then ?
\who{Patient}  nope , no issues since then .
\who{Provider}  okay , all right . and then i know that , uh , you know , you have this slightly reduced heart function , you know , your congestive heart failure . how have you been doing watching your salt intake ? i know that that's kind of been a struggle for you .
\ldots
\end{chat}
\vspace{-2em}
\end{tcolorbox}
\caption{Full chat for Figure \ref{fig:intro_example}, continued in Figure \ref{fig:example_short_chat_2}}
\label{fig:example_short_chat}
\end{figure*}

\begin{figure*}[p]
\begin{tcolorbox}[boxsep=1pt,left=2pt,right=2pt,top=0pt,bottom=0pt]
\textbf{\footnotesize Continued from \ref{fig:example_short_chat}}
\small
\vspace{-2em}
\begin{chat}

\who{Patient}  um , it's been more of a struggle recently . i've been traveling a lot . i went up to vermont , um , to go , um , explore the mountains . and along the way i stopped at , you know , mcdonald's and got two cheeseburgers . and so , i , i could be doing better . i've noticed some swelling in my , my legs . um , but nothing too extreme that where i thought i should call .
\who{Provider}  okay , all right . and any shortness of breath or problems lying flat at night , anything like that ?
\who{Patient}  no , nothing like that .
\who{Provider}  okay , all right . and then in terms of the kidney stones , i know that you had those a couple years ago , as well . any recent flare ups ? have you had any , any back pain , flank pain , anything like that ?
\who{Patient}  no , nothing like that .
\who{Provider}  okay . any blood in your urine that you've seen ?
\who{Patient}  no .
\who{Provider}  okay , all right . um , okay . well , i know that the nurse did a review of system sheet when you came in . and we've just talked a lot about your , your s- your symptoms , you know , your dizziness , your fatigue and that type of thing . anything else that i might have missed , fever chills , any nasal congestion , sore throat , cough ?
\who{Patient}  uh , i've had a little bit of nasal congestion just because with the seasons changing , i , i get seasonal allergies . but everything else has been okay ."
\ldots
\end{chat}
\vspace{-2em}
\end{tcolorbox}
\caption{Full chat for Figure \ref{fig:intro_example}}
\label{fig:example_short_chat_2}
\end{figure*}

\begin{table*}
\centering
\small
\begin{tabular}{ll|ll|rr|rr}
 &  &  &  & \multicolumn{2}{c|}{Omm. Weight} & \multicolumn{2}{c}{Omm. Count} \\
Summ. & Metric & Comp. Metric & Comp. Mean & Spear. & Pear. & Spear. & Pear. \\  \midrule
gpt4 & gpt4 & rouge1 & 0.4495 & 0.002 & 0.001 & -0.042 & -0.016 \\
gpt4 & gpt4 & rouge2 & 0.1929 & 0.016 & -0.007 & -0.046 & -0.036 \\
gpt4 & gpt4 & rougeL & 0.2652 & -0.067 & -0.082 & -0.148 & -0.129 \\
gpt4 & gpt4 & rougeLsum & 0.3630 & 0.003 & -0.041 & -0.046 & -0.057 \\
gpt4 & gpt4 & bertscore-P & 0.6684 & -0.105 & -0.038 & -0.130 & -0.043 \\
gpt4 & gpt4 & bertscore-R & 0.6368 & 0.055 & 0.027 & 0.061 & 0.058 \\
gpt4 & gpt4 & bertscore-f1 & 0.6507 & -0.044 & -0.009 & -0.061 & 0.000 \\ \midrule
gpt-3.5-t & gpt4 & rouge1 & 0.3892 & \textbf{-0.273} & \textbf{-0.239} & \textbf{-0.279} & \textbf{-0.212} \\
gpt-3.5-t & gpt4 & rouge2 & 0.1566 & \textbf{-0.253} & \textbf{-0.270} & \textbf{-0.255} & \textbf{-0.252} \\
gpt-3.5-t & gpt4 & rougeL & 0.2404 & \textbf{-0.301} & \textbf{-0.291} & \textbf{-0.314} & \textbf{-0.289} \\
gpt-3.5-t & gpt4 & rougeLsum & 0.3329 & \textbf{-0.242} & \textbf{-0.220} & \textbf{-0.244} & \textbf{-0.200} \\
gpt-3.5-t & gpt4 & bertscore-P & 0.6450 & \textbf{-0.333} & \textbf{-0.294} & \textbf{-0.338} & \textbf{-0.260} \\
gpt-3.5-t & gpt4 & bertscore-R & 0.6127 & \textbf{-0.305} & \textbf{-0.246} & \textbf{-0.299} & \textbf{-0.259} \\
gpt-3.5-t & gpt4 & bertscore-f1 & 0.6270 & \textbf{-0.338} & \textbf{-0.299} & \textbf{-0.338} & \textbf{-0.281}
\end{tabular}
\caption{Full correlation results between the omission weight and count, and all Rouge and BertScore components. The values in bold are found to be significant with a two-sided test $p<0.05$.}
\label{tab:app_corr}

\end{table*}

\begin{figure*}[p]
\begin{tcolorbox}[boxsep=1pt,left=2pt,right=2pt,top=0pt,bottom=0pt]
\textbf{\footnotesize Provider-Patient Chat}
\small
\vspace{-2em}
\begin{chat}
\ldots
\who{Provider} martha is a 50-year-old female with a past medical history significant for congestive heart failure , depression and hypertension who presents for her annual exam . so , martha , it's been a year since i've seen you . how are you doing ?
\who{Patient}  i'm doing well . i've been traveling a lot recently since things have , have gotten a bit lighter . and i got my , my vaccine , so i feel safer about traveling . i've been doing a lot of hiking . uh , went to washington last weekend to hike in northern cascades, like around the mount baker area .
\who{Provider} nice . that's great . i'm glad to hear that you're staying active , you know . i , i just love this weather . i'm so happy the summer is over . i'm definitely more of a fall person .
\who{Patient} yes , fall foliage is the best .
\who{Provider} yeah . um , so tell me , how are you doing with the congestive heart failure ? how are you doing watching your diet ? i know we've talked about watching a low sodium diet . are you doing okay with that ?
\who{Patient}  i've been doing well with that . i resisted , as much , as i could , from the tater tots , you know , the soft pretzels , the salty foods that i , i love to eat . and i've been doing a really good job .
\who{Provider} okay , all right . well , i'm glad to hear that . and you're taking your medication ?
\who{Patient}  yes .
\who{Provider} okay , good . and any symptoms like chest pains , shortness of breath , any swelling in your legs ?
\who{Patient} no , not that i've noticed .
\who{Provider} okay , all right . and then in terms of your depression , i know that we tried to stay off of medication in the past because you're on medications for your other problems . how are you doing ? and i know that you enrolled into therapy . is that helping ? or-
\who{Patient} yeah , it's been helping a lot . i've been going every week , um , for the past year since my last annual exam . and that's been really helpful for me.
\ldots
\end{chat}
\vspace{-2em}
\end{tcolorbox}

\begin{tcolorbox}[boxsep=1pt,left=2pt,right=2pt,top=0pt,bottom=0pt]
\small
\textbf{\footnotesize Subjective (Prompt \ref{prompt:summary})}

The patient, Martha, is a 50-year-old female with a history of congestive heart failure, depression, and hypertension. She presented for her annual exam and reported that she has been doing well overall. She has been traveling and hiking a lot recently, which indicates an active lifestyle. Regarding her congestive heart failure, she has been adhering to a low sodium diet and taking her medication regularly. She reported no symptoms such as chest pains, shortness of breath, or swelling in her legs. As for her depression, she has been attending therapy sessions weekly for the past year and reported that it has been helpful. She denied any feelings of wanting to harm herself or others. However, she admitted to forgetting to take her blood pressure medication occasionally, especially during stressful periods at work. She has been working long hours and late nights, which has been causing her stress and increasing her blood pressure.
\end{tcolorbox}

\caption{Example Provider-Patient chat from ACI training set.  We include  the generated Subjective.  Note that the chat and facts were truncated for length.  We include the unique fact identifiers (F + NUMBER) for reference.  For additional output, see Figure \ref{fig:example_pt2} and \ref{fig:example_pt3}.  All output was generated with GPT-4.}\label{fig:example_pt1}

\end{figure*}

\begin{figure*}

\begin{tcolorbox}[boxsep=1pt,left=2pt,right=2pt,top=0pt,bottom=0pt]
\textbf{\footnotesize Fact Extraction and Categorization (Prompt \ref{prompt:fact_categorization})}

\small
\textbf{Critical}

F1: Martha has a past medical history significant for congestive heart failure, depression and hypertension

F14: Martha is forgetting to take her blood pressure medication

F15: Martha's blood pressure goes up when work gets more stressful

\textbf{Important}

F12: Martha has been going to therapy every week for the past year

F16: Martha has been working long hours, late nights

F17: Martha has a lot of fiscal year end data that she's been having to pull

\ldots

\textbf{Other}

F0: Martha is a 50-year-old female

F2: Martha presents for her annual exam

F3: Martha has been traveling a lot recently

\ldots
\end{tcolorbox}
\begin{tcolorbox}[boxsep=1pt,left=2pt,right=2pt,top=0pt,bottom=0pt]
\small

\textbf{\footnotesize Supportive Sub-Clustering (Prompt \ref{prompt:fact_clustering})}

\textbf{Uncontrolled Hypertension}

SOCIAL DETERMINATE OF HEALTH:

\quad Work-related stress: (F15) Martha's blood pressure goes up when work gets more stressful
    
SYMPTOMS:

\quad Stress-induced hypertension: (F15) Martha's blood pressure goes up when work gets more stressful
    
TESTS: NONE
  
TREATMENTS:

\quad  Medication non-adherence: (F14) Martha is forgetting to take her blood pressure medication

\textbf{Well-managed Congestive Heart Failure}

OTHER:

\quad  Past medical history: (F1) Martha has a past medical history significant for congestive heart failure
    
SOCIAL DETERMINATE OF HEALTH: NONE

SYMPTOMS:

 \quad NONE: (F10) Martha has no symptoms like chest pains, shortness of breath, any swelling
      in her legs
      
TESTS: NONE
 
TREATMENTS:

\quad  Medication adherence: (F9) Martha is taking her medication
 
\ldots

\end{tcolorbox}
    \caption{Following from Figure \ref{fig:example_pt1}, the Supportive  and  the list and categorization of facts. Continued in Figure \ref{fig:example_pt3}}
    \label{fig:example_pt2}
\end{figure*}
\begin{figure*}
    \begin{tcolorbox}[boxsep=1pt,left=2pt,right=2pt,top=0pt,bottom=0pt]
\small

\textbf{\footnotesize Refuting Sub-Clustering }

\textbf{Uncontrolled Hypertension}

SOCIAL DETERMINATE OF HEALTH: NONE
  
SYMPTOMS:  NONE
  
TESTS: NONE
  
TREATMENTS:

\quad   Antihypertensive Medication Adherence:
  
\quad  \quad     F14: Martha is forgetting to take her blood pressure medication
    
 \quad  \quad    F9: Martha is taking her medication

\textbf{Well-managed Congestive Heart Failure}

SYMPTOMS:

\quad  Cardiac Insufficiency:
  
\quad \quad    F10: Martha has no symptoms like chest pains, shortness of breath, any swelling in her legs
    
Social Determinate of Health: NONE

TESTS: NONE

TREATMENTS: NONE

\ldots

\end{tcolorbox}

    \begin{tcolorbox}[boxsep=1pt,left=2pt,right=2pt,top=0pt,bottom=0pt]
\small

\textbf{\footnotesize Omitted Facts, Generated Explanation, and Omission Weight (Prompt \ref{prompt:omission_detection} for extraction and explanation)}

  F17: Martha has a lot of fiscal year end data that she's been having to pull
  
  \quad The summary does not mention the specific work-related stressors Martha is dealing  with. (Score : 0.5)
    
  F18: Martha has a lot of responsibility at work
  
  \quad The summary does not mention the level of responsibility Martha has at work. (Score : 0.5)
  
  F4: Martha got her vaccine
  
  \quad The summary does not mention that Martha received her vaccine. (Score: 0.1)
  
  F6: Martha went to Washington last weekend to hike in northern cascades, like around the mount baker area (Score : 0.1)
  
  \quad The summary does not specify the location of Martha's recent hiking trips. (Score : 0.1)
  
  F8: Martha has been resisting from the tater tots, the soft pretzels, the salty foods (Score : 0.5)
  
  \quad The summary does not detail the specific foods Martha has been avoiding in her  low sodium diet.

\end{tcolorbox}

    \caption{Following from Figure \ref{fig:example_pt1} and \ref{fig:example_pt2}, the Refuting Sub-clustering, and the list of missing facts.  Note that there are seemingly conflicting facts in the Refuting sub-clustering example.  However, this represents exactly what was discussed in the chat. Initially, the patient says they are taking their medication, and later says they are forgetting their blood pressure medication specifically.}
    \label{fig:example_pt3}
\end{figure*}

\begin{lstlisting}[language=promptlanguage,float=*,caption=Prompt for generating summary,label={prompt:summary}]
Below is a medical encounter between a patient and a doctor done over chat .
----
Medical Encounter
----
{{dialogue}}
----
Summary Instructions
----
Provide a summary of the medical encounter between the doctor and the patient.

Separate the note into separate sections, with divisions were inspired by the SOAP standard.
-The "Subjective" includes items taken during verbal exam and typically written in the form of chief complaint (CC), history of present illness (HPI), and past social history
DO NOT INCLUDE THE FOLLOWING SECTIONS;
-You should not include any "Objective Exam" includes content from the physical examination on the day of the visit
-You should not include any "Objective Results", which includes diagnostics taken prior to the visit, including laboratory or imaging results
-You should not include any "Assessment and Plan", which includes the doctor's diagnosis and planned tests and treatments

If there is no information for a section, please omit it.

Summary of Medical Encounter:
\end{lstlisting}

\begin{lstlisting}[language=promptlanguage,float=*,caption=Prompt for detecting fact omissions from summary,label={prompt:omission_detection}]
Instructions
-The following is a medical summary of a single medical encounter.  In addition, there is a list of facts from that same encounter.
-Acting as a medical expert who is testing medical students on their thoroughness, which facts were omitted from the summary?
-For a fact to be an omission, relevant information from the fact must be omitted.  The fact does not have to be written verbatim.
-Output the list of facts that were omitted, report the fact id, fact, and a short explanation.

--Begin Summary--
{{subjective}}
--End Summary--
--Begin Facts--
{{fact_list}}
--End Facts--

Are there any facts missing from the summary? Report the fact number, the fact, and an explanation for each.

The output should be in a json dictionary, with the following format;
{
"FACT_NUM" : ["FACT", "EXPLANATION"]
...
}
If there are no missing facts, return an empty json dictionary.

Missing facts:
\end{lstlisting}

\begin{lstlisting}[language=promptlanguage,float=*,caption=Prompt for assigning categories to each prompt,label={prompt:fact_categorization}]
You are an expert medical data labeler. You will be provided with a differential diagnosis (DDx) for a patient case and a set of medical facts describing the patient. Your task is to group these facts into 3 groups: "critical", "important", and "other". "Critical" facts are absolutely critical in order to arrive at the DDx. If this fact is not present, the DDx would be greatly altered. "Important" facts are helpful in determining the DDX, and may or may not greatly affect the DDx. "Other" facts are facts that are neither "critical" nor "important".

---Differential diagnosis (start)---
{{ddx}}
---Differential diagnosis (end)---

---Medical facts (start)---
{{facts}}
---Medical facts (end)---

Given this information, produce a numbered, ranked list of unique grouped facts.
For each category, output the category name ("Category|[CATEGORY]\n") followed by the list of facts for that category each on its own line ("[Fact_Rank]|[Fact Num]|[Fact]").

Output:
\end{lstlisting}

\begin{lstlisting}[language=promptlanguage,float=*,caption=Prompt for clustering supportive facts by diagnosis,label={prompt:fact_clustering}]
The following is a list of facts extracted from a medical encounter.


Your role is to select which  positive fact(s) support each diagnosis.
Therefore, only report pertinent positives which support each diagnosis.  Do not report supportive results that negate the diagnosis, or any other type of fact.


A fact can occur in multiple diagnoses.

The classifications should be in reference to this differential diagnosis;
{{ddx}}

Facts:
{{facts}}


Output the results in a json dictionary, such as;
{
"DIAGNOSIS 1" : {"FACT_NUM" : "EXPLANATIION" ...}
...
}
If a diagnosis has no facts, output an empty array.

Clusters:
\end{lstlisting}

\begin{lstlisting}[language=promptlanguage,float=*,caption=Prompt for truncating dialogue,label={prompt:dialogue_truncation}]
The following is a patient-doctor dialogue.


{{dialogue}}

Consider the conversation in the frame of a SOAP medical note framework.
We want to include all dialogue lines that contain information that might be relevant to the subjective.
This includes;
-Chief Complaint
-History of Present Illness
--This includes questions about the patient's current health status.
-Past medical history
--The includes any discussion of previously diagnosed medical issues.
This does not include;
-Physical exam
-Laboratory Results
-New diagnoses made by the provider in this conversation
-Assessment or care plan
Return the last line of the conversation that collects this information.

The conversation begins with line number 0.
Output the entire relevant line in a valid json dictionary formatted as follows;
{
[LINE_NUM] : [MSG]
}
Where [LINE_NUM] is a valid integer, and [MSG] is the relevant message.


Output:
\end{lstlisting}

\end{document}